\pdfoutput=1
\PassOptionsToPackage{dvipsnames}{xcolor}
\documentclass[11pt]{article}

\usepackage{acl}

\usepackage{times}
\usepackage{latexsym}

\usepackage[T1]{fontenc}

\usepackage[utf8]{inputenc}
\usepackage[most]{tcolorbox}
\usepackage{microtype}
\usepackage{amsfonts} 
\usepackage{pifont}
\usepackage{cancel}
\usepackage{soul}
\usepackage{algorithm}
\usepackage{algorithmic}
\usepackage{booktabs}
\usepackage{xspace}
\usepackage{fontawesome}
\usepackage{mathrsfs}
\usepackage{multirow}
\usepackage{amsmath}
\usepackage{colortbl}
\usepackage{subfig}
\usepackage{graphicx}
\usepackage{tabularx}
\usepackage{arydshln}
\usepackage{makecell}
\usepackage{CJKutf8}

\usepackage{caption}       

\usepackage[dvipsnames]{xcolor}

\newcolumntype{M}{>{\columncolor{BrickRed!9}}c}      
\newcolumntype{I}{>{\columncolor{YellowGreen!12}}c}    
\newcolumntype{S}{>{\columncolor{Aquamarine!12}}c}   

\definecolor{bleudefrance}{rgb}{0.19, 0.55, 0.91}
\definecolor{yes}{RGB}{239,211,69}
\definecolor{carminered}{rgb}{1.0, 0.0, 0.22}
\definecolor{crimsonglory}{rgb}{0.75, 0.0, 0.2}

\definecolor{err}{RGB}{255,0,128}
\definecolor{corr}{RGB}{65,105,225}
\newcommand{\erritalic}[1]{\textcolor{err}{\textit{#1}}}
\newcommand{\corritalic}[1]{\textcolor{corr}{\textit{#1}}}

\title{Beyond Profile: From Surface-Level Facts to\\ Deep Persona Simulation in LLMs}

\author{
Zixiao Wang\textsuperscript{1} ~~~Duzhen Zhang\textsuperscript{1} ~~~Ishita Agrawal\textsuperscript{1}\\
\textbf{\large Shen Gao\textsuperscript{2}} ~~~
\textbf{\large Le Song\textsuperscript{1}} ~~~
\textbf{\large Xiuying Chen\textsuperscript{1}\thanks{~~Corresponding Author.}~}
\\
\textsuperscript{1}Mohamed bin Zayed University of Artificial Intelligence, \\
\textsuperscript{2}Shandong University \\
\texttt{\{zixiao.wang,duzhen.zhang,xiuying.chen\}@mbzuai.ac.ae}\\
}

\begin{document}
\maketitle
\begin{abstract}
Previous approaches to persona simulation large language models
(LLMs) have typically relied on learning basic biographical information, or using limited role-play dialogue datasets to capture a character’s responses.
However, a holistic representation of an individual goes beyond surface-level facts or conversations to deeper thoughts and thinking.
In this work, we introduce CharacterBot, a model designed to replicate both the linguistic patterns and distinctive thought patterns as manifested in the textual works of a character.
Using Lu Xun, a renowned Chinese writer as a case study, we propose four training tasks derived from his 17 essay collections.
These include a pre-training task focused on mastering external linguistic structures and knowledge, as well as three fine-tuning tasks: multiple-choice question answering, generative question answering, and style transfer, each aligning the LLM with Lu Xun’s internal ideation and writing style. 
To optimize learning across these tasks, we introduce a CharLoRA parameter updating mechanism, where a general linguistic style expert collaborates with other task-specific experts to better study both the language style and the understanding of deeper thoughts.
We evaluate CharacterBot on three tasks for linguistic accuracy and opinion comprehension, demonstrating that it significantly outperforms the baselines on our adapted metrics.
We hope this work inspires future research on deep character persona simulation LLMs {\color{black}while considering the importance of ethical standards}: \faGithub  \href{https://github.com/zxwang63/characterbot}{CharacterBot}.
\end{abstract}

\section{Introduction}

\begin{figure}[t]
\centering
\includegraphics[scale=0.75]{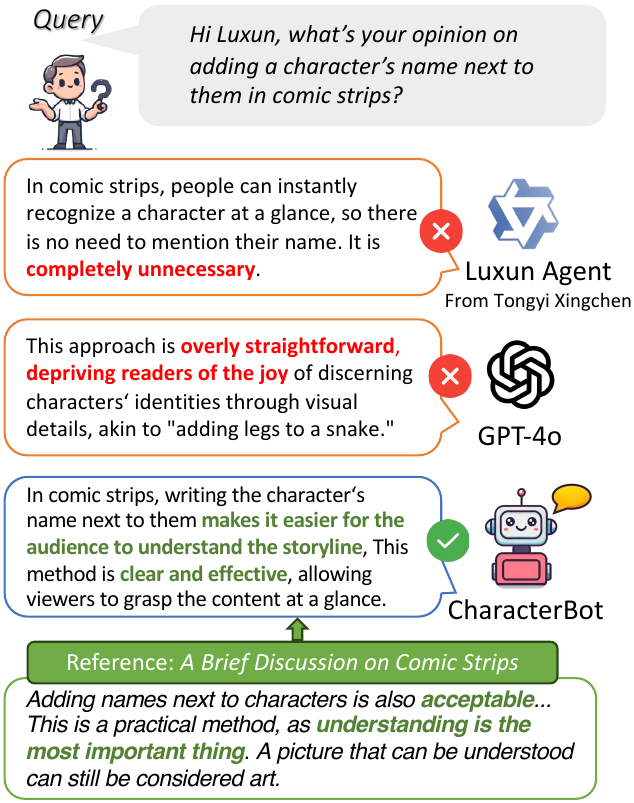}
\caption{
Comparison of CharacterBot and other models in responding to Lu Xun-related questions.
}
\label{fig:compare}
\end{figure}

Developing language models that simulate character personas by embodying specific individuals' personalities and worldviews has been a longstanding objective in NLP. 
Researchers in this area have explored various approaches. 
Some models are fine-tuned to memorize essential \textit{profile information}, such as birthdate, occupation, and other background traits~\cite{shao2023character, wang2023rolellm, bai2024baijia}. 
Others leverage \textit{conversational data} from novels, screenplays, and multimedia content to adapt language models through dialogue-based fine-tuning~\cite{li2023chatharuhi, zhang2024thinking, tu2024charactereval}. 
Additionally, prompt engineering techniques dynamically inject \textit{profile-specific information} during generation, enabling models to emulate character traits without extensive fine-tuning~\cite{han2022meet, shanahan2023role, tu2023characterchat}.

As shown in the related works above, current methods often oversimplify persona representation, limiting it to superficial dialogues or basic profile descriptors.
We argue that \textit{robust persona simulation must move beyond basic attributes and narrow conversational patterns} to incorporate deeper aspects of human identity, such as an individual's worldview, ethical frameworks, context-dependent viewpoints, and foundational beliefs.
Echoing Virginia Woolf's idea that ``books are the mirrors of the soul,'' we propose a new paradigm: deep persona simulation based on a writer's works, which inherently reflect their beliefs, insights, and reactions to diverse themes and topics, thus capturing their essence more authentically.

In this work, we explore the problem of character persona simulation using Lu Xun as an exemplar—a renowned Chinese writer known for his critical essays on sociocultural issues. We develop CharacterBot to model his persona, as shown in Figure~\ref{fig:compare}.
We curated a corpus comprising 17 essay collections, including 638 individual works containing titles, full texts, and segmented passages. 
To enable a comprehensive acquisition of both stylistic patterns and ideological depth, we implemented four distinct tasks. 
First, next-token prediction pre-training combined with authorial perspective reframing facilitates mastery of Lu Xun's linguistic style and cognitive frameworks. 
Subsequently, in the fine-tuning stage, the model learns to: 
1) resolve multiple-choice questions testing comprehension of authorial viewpoints,
2) extract core ideological propositions from textual segments, and
3) perform style transfer by transforming styleless text inputs into outputs that emulate the author's distinctive stylistic patterns.
{\color{black}
These tasks are designed to prompt the model to generate responses that align with the ideological stances or common arguments found within Lu Xun's essays.}

For model design, we develop a CharLoRA framework that extends LoRA~\cite{hulora} by optimizing knowledge integration specifically for persona simulation. 
Unlike standard LoRA, which primarily focuses on adapting models to specific tasks via low-rank updates, CharLoRA introduces a structured decomposition: a general matrix pair \(\mathbf{A}_{\text{pt}}\) and \(\mathbf{B}_i\) that jointly capture both linguistic foundations and cognitive patterns. 
While task-specific \(\mathbf{B}_i\) matrices specialize in handling task-relevant persona patterns, the shared \(\mathbf{A}_{\text{pt}}\) matrix enables cross-task knowledge synthesis.
This novel design not only preserves persona consistency across tasks but also enhances deep persona representation through multi-objective learning, distinguishing CharLoRA from conventional LoRA approaches.
Training data was generated using GPT-4o with carefully designed prompts and rigorously validated by humans to ensure quality.

Since no established metrics exist for evaluating personalized LLMs with both linguistic depth and philosophical insight, we propose novel evaluation criteria and compare our model against latest role-playing models. 
CharacterBot shows significant improvements in both style and ideological depth. Additionally, we conduct a human evaluation where participants rank outputs from our model and baselines against reference texts. 
Results confirm that our model consistently outperforms baselines in generating contextually accurate and stylistically aligned outputs.

{\color{black}
While a high-fidelity simulation of a writer's persona offers exciting possibilities, it also brings forth ethical responsibilities. 
These include ensuring that the simulation is respectful, does not misrepresent the individual's core beliefs in harmful ways, and that its outputs are clearly distinguishable from authentic works, especially when dealing with historical figures.
}

Our main contributions are as follows:  
(1) We introduce a novel framework for character persona simulation in LLMs that captures both linguistic style and deep ideological perspectives beyond superficial profile memorization.
(2) We design a multitask architecture with CharLoRA that combines general linguistic knowledge with task-specific persona modeling through tasks like question answering and style transfer.
(3) Experiments show that CharacterBot outperforms baselines in linguistic accuracy, style preservation, and opinion comprehension, simulating complex character personas.

\section{Related Work}

\paragraph{Character Persona Simulation.}
Character persona simulation refers to the task of assigning personality traits to virtual agents in order to facilitate realistic and contextually coherent interactions \cite{tseng2024two, liu2024tiny}. 
In recent years, this task has received widespread attention, particularly in approaches leveraging LLMs. 
With their strong capabilities in natural language understanding and generation \cite{song2024hazards, zhang2024mm, li2025system, zheng2025lifelong}, LLMs have become a key tool for simulating diverse personas. 
Currently, two main approaches are employed: prompt-based methods and fine-tuning-based methods.
Prompt-based methods integrate character settings directly into prompts. 
For instance, \citet{shanahan2023role} introduce a role-playing mechanism through prompts, while \citet{tu2023characterchat} construct a dialogue system using the MBTI personality framework. 
\citet{agatsuma2024building} simulate patient personas to train nursing students in health guidance. 
However, prompt engineering often fails to capture intricate traits and dynamic behaviors.
Fine-tuning-based methods address these limitations by training pre-trained LLMs. \citet{shao2023character} fine-tune models using character profiles and an experience upload strategy. 
\citet{lu2024large} extract character attributes, background details, and dialogue features from knowledge bases to fine-tune models. 
\citet{park2024enhancing} further refine role-specific behaviors through trait extraction from novel summaries. 
Despite these advancements, fine-tuning often produces models that mimic surface-level style but miss deeper traits.

\paragraph{Style Transfer.}  
Style transfer for text involves altering the stylistic attributes of a source text while preserving its core meaning \cite{li2023stylized}.
\citet{reif2022recipe} introduce an Augmented Zero-Shot Learning method, which leverages LLMs to achieve versatile text-style transformations without requiring task-specific training.
\citet{pu2023chatgpt} evaluate the performance of ChatGPT in sentence style transfer by comparing ChatGPT generated texts with those created by humans.
\citet{zhang2024distilling} explore a novel approach to style transfer by integrating the Chain-of-Thought reasoning capabilities of LLMs with knowledge distillation techniques.
However, these methods predominantly focus on linguistic style without engaging with deeper ideological dimensions.

\paragraph{Parameter-Efficient Fine-Tuning.}
Parameter-Efficient Fine-Tuning (PEFT) techniques are crucial for adapting large pre-trained models efficiently, especially in the context of LLMs \cite{xu2023parameter, ding2023parameter, chen2024flexible}. Prefix Tuning optimizes learnable prefixes added to input embeddings or intermediate representations \cite{li2021prefix}, while Prompt Tuning adjusts learnable prompt embeddings to guide downstream task adaptation \cite{lester2021power}. Among PEFT methods, LoRA is widely used, introducing low-rank matrices to fine-tune models efficiently \cite{hulora}. Its variant, QLoRA, further improves efficiency through quantization \cite{dettmers2024qlora}.
Recent innovations have improved LoRA’s domain-specific adaptability. \citet{zhao2024loraretriever} propose a dynamic mechanism for retrieving and combining LoRA modules in mixed task scenarios. \citet{tan2024personalized} enable efficient personalized LLMs by sharing partial PEFT parameters across users, and \citet{zhao2024merging} introduce a flexible LoRA merging strategy suited for multi-task learning without additional training. In our work, we adapt LoRA for deep persona simulation.

\begin{figure*}[t]
    \centering
    \includegraphics[width=0.95\linewidth]{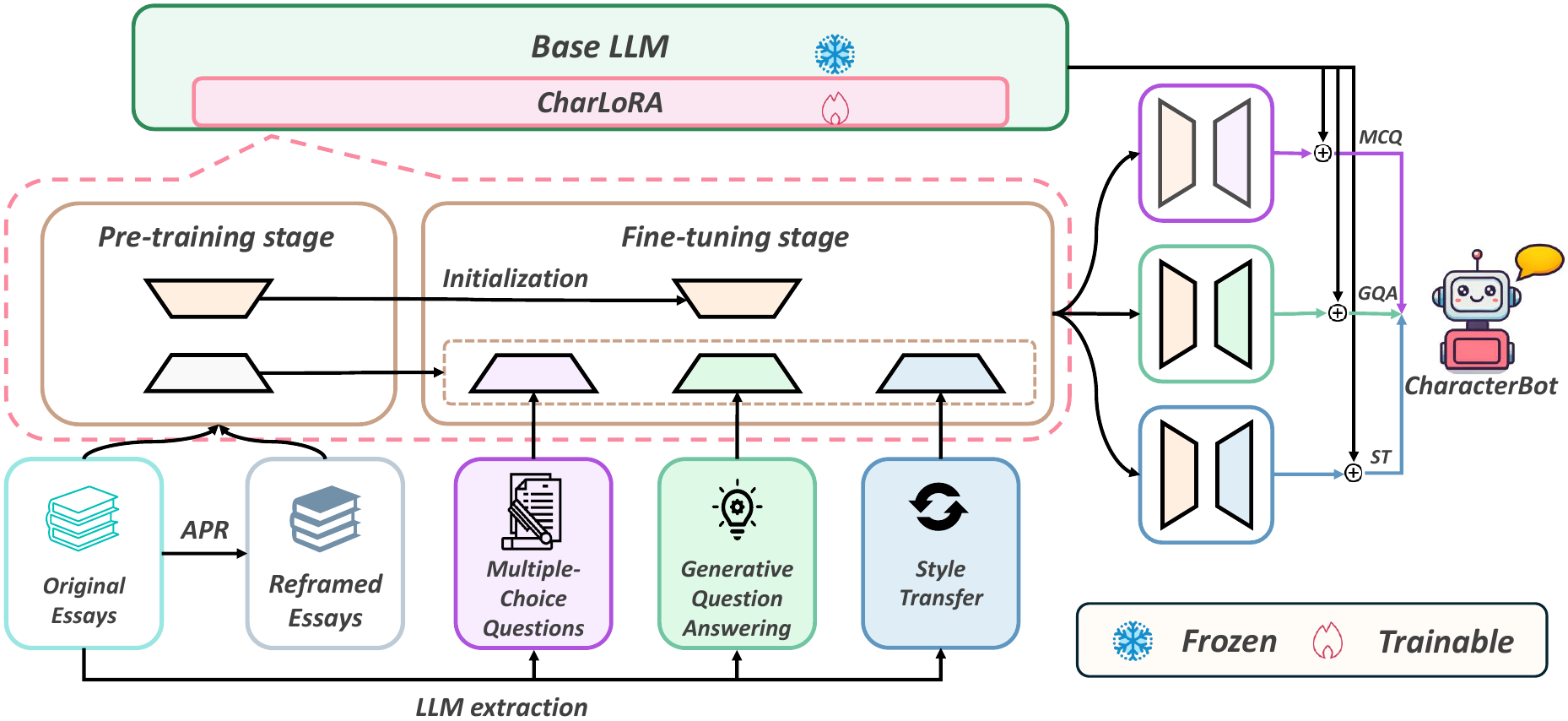}
    \caption{In pre-training, reframed essays from Authorial Perspective Reframing (APR) train the base model. In fine-tuning, multiple-choice question answering (MCQ), generative question answering (GQA), and style transfer (ST) refine their modules within CharLoRA to align with the target persona.}
    \label{fig:method}
\end{figure*}

\section{Problem Formulation}  

{\color{black}
We formally define the task of deep persona simulation as the generation of character-specific responses and stylistic outputs that align with the inferred cognitive and linguistic patterns of a target individual, primarily derived from their textual corpus. 
The depth refers to capturing nuances beyond surface-level facts, aiming for a consistent representation of style, tone, and recurring themes, rather than an ontological claim about replicating an inner self.
}

A single model \( M \) is designed to perform multiple tasks, which together assess its ability to understand and generate character-specific responses.

The first task is Multiple-Choice Question Answering, where the model \( M \) is given a question \( Q \) and a set of candidate answers \( \{A_1, A_2, \dots, A_n\} \). The model predicts the correct answer \( A^* \) based on its understanding of the character’s perspective:  
\[
A^* = \arg \max_{A_i} M(Q, \{A_1, ..., A_n\}; \theta),
\]  
where \( \theta \) denotes the model’s parameters.

The second task is Generative Question Answering, where the model \( M \) generates an appropriate answer \( A \) given a question \( Q \), reflecting the character’s knowledge, opinions, or personality:  
\[
A = M(Q; \theta).
\]  
The final task is Style Transfer, where the model \( M \) rewrites an input text \( T_{\text{input}} \) into an output text \( T_{\text{output}} \) that aligns with a specified persona style \( S \):  
\[
T_{\text{output}} = M(T_{\text{input}}, S; \theta).
\]  

\section{Method}

As shown in Figure~\ref{fig:method}, our CharacterBot is first pre-trained and then fine-tuned on three downstream tasks.
In the following sections, we introduce the task-specific objectives and adapted training datasets for each stage, followed by a detailed explanation of our customized CharLoRA.

\subsection{Pre-training}

Although conventional LLMs are typically trained on books that may include the works of Lu Xun, they are fine-tuned on broad-coverage corpora that, while fostering general linguistic competence, lack the domain-specific precision necessary to replicate nuanced authorial traits.
To address this gap, we specifically pre-train the LLM on Lu Xun’s corpus to capture his distinctive narrative style and linguistic patterns.

\paragraph{Authorial Perspective Reframing (APR).}
To enhance the model’s understanding of authorial perspectives, we introduce {\color{black}Authorial Perspective Reframing (APR)}, a pre-training technique that aligns textual viewpoints with their original intellectual contexts by transforming first-person narratives into third-person perspectives.
{\color{black}The core idea is to render implicit perspectives explicit by incorporating attribution markers such as “the author argues” or “the author critiques,” thereby establishing a direct link between each statement and its originator.}

This transformation clarifies attribution by explicitly linking content to its creator, thereby reducing ambiguity. 
It also improves the model’s ability to distinguish between described concepts and the author’s stance, while reinforcing the connection between viewpoints and their originators.
{\color{black}And it reinforces ideological ownership, facilitating the model’s association of specific viewpoints with particular authors.}

{\color{black}
To automate this transformation process, we utilize GPT-4o-mini with a carefully designed prompt, as detailed in Appendix~\ref{sec:prompts}, Table~\ref{tab:prompt4_paraphrase} to convert raw essay texts into their APR versions. The model systematically incorporates authorial cues while preserving the original semantic content.
For instance, consider the following transformation based on an essay by Lu Xun:
}
\begin{tcolorbox}[colback=gray!10, left=1mm, right=1mm, top=1mm, bottom=1mm] [Original Essay] Comparison is the best thing. Before understanding phonetic scripts, people would find it difficult to realize the challenges of pictographic characters... \end{tcolorbox}
\begin{tcolorbox}[colback=gray!10, left=1mm, right=1mm, top=1mm, bottom=1mm] [APR Processed] \textbf{Lu Xun discusses} the importance of comparison in his essay. \textbf{He points out} that before understanding phonetic scripts, it is hard for people to realize the challenges of pictographic characters... \end{tcolorbox}
{\color{black}
By externalizing the author’s voice, APR facilitates a more distinct semantic structure, thereby enhancing the model’s perspective alignment and contextual comprehension during the pre-training stage.
}
\subsection{Fine-tuning}

Following pre-training, we fine-tune the LLM on three tasks: Multiple Choice Questions, Generative Question Answering and Style Transfer to align it with the cognitive patterns, style, and ideology of the target character. For examples of three tasks, refer to Appendix~\ref{sec:data-examples}.

\paragraph{Task 1: Multiple Choice Questions.}
The motivation for this task comes from the common use of multiple choice questions in reading comprehension exams in various languages, making it an effective way to assess whether the bot truly understands the author’s ideas within the essay.
Our first fine-tuning task involves multiple-choice answering, where each question, derived from an essay, provides four answer options with one correct answer.
To ensure alignment with the essay content, the dataset generation process follows two key constraints: The questions should be directly grounded in the essay, avoiding any extrapolation, and they must use second-person framing by addressing the character as ``you.''

\paragraph{Task 2: Generative Question Answering.}  
Similarly, another common type of reading comprehension question involves asking respondents to answer based on their understanding of an article.  
In this task, the generative question answering creates pairs of questions and answers that investigate the underlying arguments of the character.  
The questions are framed as ideological probes in the second person, while the answers reflect the rhetorical and ideological patterns of the character.  
For each essay, three pairs are created, following two principles: semantic fidelity, ensuring that the answers derive solely from the essay, and stylistic consistency, preserving the unique lexical and syntactic characteristics of the character.

\paragraph{Task 3: Style Transfer.}  
The final training task focuses on style transfer, commonly used in personalization tasks where the model rewrites sentences to follow a specific style. For each essay, three sentences that are representative of the character’s style are extracted using an advanced language model. 
These sentences are then rewritten into a neutral, styleless form that preserves their original meaning, argumentative intent, and emotional tone while removing distinct stylistic features.  
This parallel structure enables the model to learn how to rewrite text in the author’s style while maintaining the intended meaning and ideological consistency.

\subsection{CharLoRA}  

To optimize computational and storage efficiency while ensuring high persona fidelity, we introduce CharLoRA, an adaptation of the LoRA method. 

\paragraph{LoRA Fundamentals.}  
For a frozen pre-trained weight matrix \( W_0 \in \mathbb{R}^{d \times k} \) in any linear layer, LoRA injects trainable low-rank matrices \( \mathbf{B} \in \mathbb{R}^{d \times r} \) and \( \mathbf{A} \in \mathbb{R}^{r \times k} \), with \( r \ll \min(d,k) \), to approximate parameter updates:  
\[
W_0 + \Delta W = W_0 + \mathbf{B}\mathbf{A},  
\]   
Given input activations \( \mathbf{x} \in \mathbb{R}^k \), the modified forward pass becomes:  
\[
\mathbf{h}' = W_0\mathbf{x} + \mathbf{B}\mathbf{A}\mathbf{x}.  
\]  
During optimization, \( W_0 \) remains fixed, while \( \mathbf{A} \) and \( \mathbf{B} \) accumulate gradients - reducing trainable parameters versus full fine-tuning.  

\paragraph{CharLoRA in Pre-Training Stage.}
In the pre-training stage, the primary goal of CharLoRA is to inject persona-specific knowledge into a base language model without disrupting its general pre-trained knowledge. To achieve this, we leverage low-rank matrix adaptations through LoRA while maintaining a separation between the original model parameters and persona-specific updates.  

The process begins with the reframed versions of the original texts using APR. 
CharLoRA applies a low-rank adaptation to the weight matrix of the frozen base model \( W_0 \), introducing an update matrix \( \Delta W_{\text{pt}} \) defined as:  
\[
\mathbf{h}'_{\text{pt}} = W_0 \mathbf{x} + \Delta W_{\text{pt}} \mathbf{x} = W_0 \mathbf{x} + \mathbf{B}_{\text{pt}} \mathbf{A}_{\text{pt}} \mathbf{x},
\]  
where \( \mathbf{B}_{\text{pt}} \in \mathbb{R}^{d \times r} \) and \( \mathbf{A}_{\text{pt}} \in \mathbb{R}^{r \times k} \) are low-rank matrices with \( r \ll \min(d, k) \).  

During training, the original weight matrix \( W_0 \) remains frozen, while only the low-rank matrices \( \mathbf{B}_{\text{pt}} \) and \( \mathbf{A}_{\text{pt}} \) are optimized. The update \( \mathbf{B}_{\text{pt}} \mathbf{A}_{\text{pt}} \) encodes fine-grained persona-specific traits, such as writing style, tone, and ideological perspectives. 
For instance, when pre-training on Lu Xun’s original essays and their reframed versions, this update effectively captures his distinctive voice and literary style. 
By decoupling the base model from low-rank adaptations, CharLoRA ensures that new persona-specific knowledge is efficiently integrated while preserving the general knowledge of the original language model.

\paragraph{CharLoRA in the Fine-Tuning Stage.}  
Building on the persona-specific knowledge encoded during pre-training, CharLoRA is designed to handle diverse downstream tasks while preserving the learned persona consistency across tasks. 
However, each task has unique demands, such as varying output formats and contextual focus, which necessitate task-specific adaptations. 
To address this, CharLoRA adopts a hybrid parameter-sharing strategy by decoupling shared persona knowledge from task-specific updates.

\begin{table*}[t]
    \centering
    \small
    \begin{tabularx}{\linewidth}{>{\centering\arraybackslash}X M I I S S S}
    \toprule
    \multirow{2}{*}{\textbf{Model}}
    & \multicolumn{1}{c}{\textbf{M-C Questions}} 
    & \multicolumn{2}{c}{\textbf{Generative Question Answering}} 
    & \multicolumn{3}{c}{\textbf{Style Transfer}}               
    \\ 
    \cmidrule(lr){2-2}
    \cmidrule(lr){3-4}
    \cmidrule(lr){5-7}
    & \textbf{Accuracy} 
    & \textbf{Content Score} 
    & \textbf{Style Score} 
    & \textbf{BLEU} 
    & \textbf{ROUGE-1} 
    & \textbf{Style Matching} 
    \\ 
    \midrule
    Llama3.1-8B         & 0.614    & 2.370            & 1.354            & 0.113          & 0.264          & 0.267          \\
    Qwen2.5-7B           & 0.787    & 2.828            & 2.818            & 0.115          & 0.233          & 0.456          \\
    GPT-4o               & 0.734    & \textbf{3.214}   & 2.542            & 0.088          & 0.196          & 0.471          \\
    CharacterGLM-6B      & 0.073    & 1.984            & 1.729            & 0.017          & 0.084          & 0.351          \\
    Baichuan-NPC-Turbo   & 0.568    & 2.620            & 2.052            & 0.124          & 0.185          & 0.518          \\
    Tongyi Xingchen      & \underline{0.788}    & 3.172            & \underline{2.823} & 0.101          & 0.187          & \underline{0.534}          \\
    LuXun-GPT            & -        & -                & -                & \underline{0.127}          & \underline{0.283}          & 0.387          \\
    Ours                 & \textbf{0.880} 
                         & \textbf{3.214} 
                         & \textbf{2.885}
                         & \textbf{0.293} 
                         & \textbf{0.410} 
                         & \textbf{0.937} 
    \\ 
    \bottomrule
    \end{tabularx}
    \caption{Results on Multiple-Choice Questions, Generative Question Answering, and Style Transfer. 
    M-C Questions denotes Multiple-Choice Questions.
    Note that LuXun-GPT is solely designed for the style transfer task, so only style transfer results are available. 
    Higher values indicate better performance.
    The best results are shown in bold, and the second best scores are underlined.}
    \label{tab:experiment}
\end{table*}
The shared low-rank matrix \( \mathbf{A}_{\text{pt}} \), initialized during pre-training, encodes core persona attributes like writing style and themes. For fine-tuning, CharLoRA replicates the pre-trained matrix \( \mathbf{B}_{\text{pt}} \) into task-specific matrices \( \mathbf{B}_i \), allowing customization for each downstream task \( i \) (e.g. multiple choice questions, generative QA, or style transfer) while retaining global persona knowledge. The forward pass for task \( i \) is defined as:
\[
\mathbf{h}_i = W_0 \mathbf{x} + \Delta W_i \mathbf{x} = W_0 \mathbf{x} + \mathbf{B}_i \mathbf{A}_{\text{pt}} \mathbf{x}.
\]  

During fine-tuning, only \( \mathbf{B}_i \) and the shared \( \mathbf{A}_{\text{pt}} \) are updated for the active task \( i \), while the matrices corresponding to other tasks \( \mathbf{B}_j \) (\( j \neq i \)) remain frozen to prevent interference. 
By maintaining \( \mathbf{A}_{\text{pt}} \) as a shared cross-task component and allowing each task-specific \( \mathbf{B}_i \) to capture task-specific details, CharLoRA achieves an optimal balance between efficiency and adaptability. For example, \( \mathbf{A}_{\text{pt}} \) ensures that Lu Xun’s characteristic voice is preserved throughout the tasks, while each \( \mathbf{B}_i \) allows nuanced adaptations, such as adjusting the tone for answering questions or generating stylistically consistent content in different contexts. This design allows CharLoRA to deliver robust performance in multi-task learning scenarios while maintaining persona consistency.

\section{Experiments}
\label{sec:experiment}
\subsection{Datasets}

In this work, we examined the writings of the renowned Chinese author Lu Xun as a case study, utilizing 17 of his essay collections comprising 638 essays. 
The pre-training stage used both the original essay texts and their transformed versions processed through Authorial Perspective Reframing (APR), which were employed in a next-token prediction task. 
The pre-training dataset, sourced from Wikisource, includes major essay collections such as \textit{False Freedom}, \textit{Grave}, and \textit{Hot Wind}. 
These collections, which vividly convey Lu Xun’s ideological perspectives, provide essential material for capturing his unique narrative style. 
A complete list of the collections can be found in the Appendix~\ref{sec:essay-lists}.
For the fine-tuning stage, we constructed three datasets derived from Lu Xun’s essays using OpenAI’s GPT-4o API. 
Each dataset was randomly split into training (85\%), validation (5\%), and testing (10\%) sets to ensure unbiased evaluation.
While brief descriptions of these datasets were provided during the task explanations, Table~\ref{tab:dataset} summarizes key statistics for further reference.

\begin{table}[htbp]
\centering
\small
\resizebox{0.482\textwidth}{!}{
\begin{tabular}{@{}ccc@{}}
\toprule
\textbf{\begin{tabular}[c]{@{}c@{}}Essay\\ Collections\end{tabular}} & \textbf{\begin{tabular}[c]{@{}c@{}}Total\\ Essays\end{tabular}} & \textbf{Task Type (Count)} \\ \midrule
\multirow{3}{*}{17} & \multirow{3}{*}{638} & Multiple-Choice Questions (1914) \\
                    &                      & Generative Question Answering (1914) \\
                    &                      & Style Transfer (1907) \\ \bottomrule
\end{tabular}
}
\caption{Summary of Lu Xun’s essay collections and the three fine-tuning tasks, showing the task types and their respective instance counts.}
\label{tab:dataset}
\end{table}

\subsection{Baselines}

To evaluate the performance of our proposed model, we conducted benchmark evaluations of advanced open-source and proprietary conversational LLMs to assess their performance:
(1) Llama3.1-8B \cite{dubey2024llama}: A multilingual LLM by Meta for diverse language tasks.
(2) Qwen2.5-7B \cite{yang2024qwen2}: An LLM by Alibaba, optimized for advanced Chinese processing.
(3) GPT-4o: A multimodal LLM by OpenAI with state-of-the-art cross-modal capabilities.
Additionally, we included models explicitly designed for role-playing applications:
(4) CharacterGLM-6B \cite{zhou2023characterglm}: A role-based dialogue model built upon the ChatGLM series.
(5) \href{https://npc.baichuan-ai.com}{Baichuan-NPC-Turbo}: An advanced LLM developed by Baichuan Intelligence, focusing on dynamic role-playing scenarios.
(6) \href{https://tongyi.aliyun.com/xingchen}{Tongyi Xingchen}: A personalized role-dialogue platform launched by Alibaba Cloud, built on the Qwen LLM framework.
Beyond general-purpose and role-playing models, we also evaluated a specialized model for text style transfer:
(7) \href{https://github.com/Suffoquer-fang/LuXun-GPT}{LuXun-GPT}: An open-source project designed to transform input into the linguistic style of Lu Xun.

\subsection{Implementation Details}
The experiments were conducted using PyTorch on an NVIDIA A100 GPU. Since Lu Xun’s writings are in Chinese, we selected the Qwen2.5-7B-Instruct model, an LLM optimized for the Chinese language, as the base model. 
During the pre-training stage, CharLoRA was applied with a LoRA rank of 64 to introduce low-rank adaptations. 
The training configuration included a token cutoff length of 2048, a learning rate of \(5.0 \times 10^{-5}\), three training epochs, a batch size of 4. 
For the fine-tuning stage, the LoRA rank was kept at 64, and the same learning rate, batch size, and validation set size were used. 
The token cutoff length was reduced to 1024, and fine-tuning was performed over three epochs.

\subsection{Evaluation Metrics}
We design classic and advanced aspect-based metrics to evaluate performance on different tasks.

For the multiple-choice questions task, performance is measured using accuracy, calculated as the number of correct answers divided by the total number of test cases.
For the generative question answering task, two metrics are used to evaluate performance: the content score and the style score. The content score assesses how well the generated answers align with the core ideological ideas on a scale from 1 to 5, while the style score evaluates the adherence to the author’s linguistic patterns, also on a scale from 1 to 5. 
For the style transfer task, we compared the model-generated sentences for each test example with sentences from Lu Xun's original work. The evaluation used BLEU \cite{papineni2002bleu}, ROUGE-1 \cite{lin2004rouge}, and the style matching score, which measures the degree to which the generated texts reflect the author’s distinctive style.

\begin{figure*}[htb]
    \centering
    \includegraphics[width=1\linewidth]{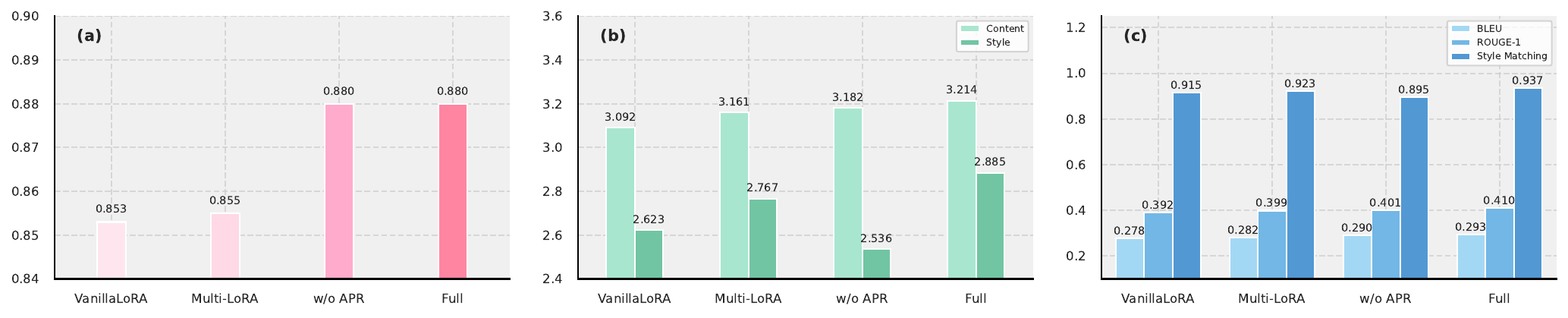}
    \caption{Ablation study results: (a) multiple-choice questions, (b) generative question answering, and (c) style transfer. The full architecture outperforms ablated versions, demonstrating the contributions of CharLoRA and APR to task performance.}
    \label{fig:ablation}
\end{figure*}

The evaluation process is conducted using DeepSeek-V3 \cite{liu2024deepseek}, with its strong capabilities of Chinese LLMs, and two human annotators, both native speakers with extensive literary expertise: one with a PhD and the other a PhD candidate.
Detailed scoring guidelines and prompts are provided in the Appendix~\ref{sec:prompts}.

\subsection{Main Results}
We show the performance of our model in Table~\ref{tab:experiment}. \textit{Our model achieves the highest accuracy in multiple-choice questions, excelling at capturing complex ideological nuances.} It achieves an accuracy of 0.880, outperforming Tongyi Xingchen (0.788) and Qwen2.5-7B (0.787). This highlights its effectiveness in understanding personality-driven narratives. Tongyi Xingchen and Qwen2.5-7B also outperform models like Llama3.1-8B (0.614) due to their extensive use of Chinese-language data, emphasizing the importance of linguistic alignment. Further pre-training and fine-tuning enhance Qwen2.5-7B’s grasp of Lu Xun’s ideological nuances.

For the generative question answering task, \textit{our model excels in preserving both content and stylistic fidelity, outperforming key baselines in style.} It achieves a content score of 3.214 and a style score of 2.885, surpassing GPT-4o’s 2.542 and Tongyi Xingchen’s 2.823. This demonstrates its superior ability to reflect the author’s style. Additionally, the correlation between content scores and multiple-choice accuracy shows that improved character understanding enhances stylistic alignment.

In the style transfer task, \textit{our model excels in capturing and reproducing the author’s style while maintaining content integrity}. It outperforms baselines with scores of 0.293 (BLEU), 0.410 (ROUGE-1), and 0.937 (style matching), demonstrating strong stylistic fidelity and textual coherence. These results underscore the importance of integrating ideological and stylistic frameworks for authentic reproduction, surpassing LuXun-GPT in overall performance.

\section{Analysis and Discussion}

\subsection{Ablation Study}
We assess the impact of individual components by evaluating ablated versions across three tasks. As shown in Figure~\ref{fig:ablation}, the full architecture achieves the best performance, confirming that both CharLoRA and APR are essential for improving downstream tasks.

{\color{black}
To analyze the contributions of CharLoRA's specific architectural choices, we examine simpler LoRA configurations. The VanillaLoRA setup, representing a more standard LoRA implementation without the enhancements of CharLoRA, shows noticeable performance drops across all metrics compared to the full CharLoRA model. Specifically, it reduces the content score (Figure~\ref{fig:ablation}(a)), significantly degrades the style score from human evaluation (Figure~\ref{fig:ablation}(b)), and lowers the automatic style matching, BLEU, and ROUGE-1 scores (Figure~\ref{fig:ablation}(c)). The Multi-LoRA variant performs slightly better than VanillaLoRA across these metrics but still lags behind the full CharLoRA model. These results underscore that the specific architectural components within CharLoRA are crucial for robust style representation and maintaining content quality.
}

Similarly, excluding APR (w/o APR) weakens the model’s ability to process stylistic information, causing a sharp decline in style score to 2.536 (Figure~\ref{fig:ablation}(b)) and style matching score to 0.895 (Figure~\ref{fig:ablation}(c)). 
This highlights APR’s crucial role in decoupling narrative perspective from content through third-person reframing.

These results demonstrate that CharLoRA and APR work together to enhance style-sensitive tasks while preserving content fidelity, reinforcing the effectiveness of our design choices.

\begin{table}[!b]
\centering
\scriptsize
\begin{tabular}{@{}cccc@{}}
\toprule
\multirow{2}{*}{\textbf{Model}} & \multicolumn{2}{c}{\textbf{Generative Question Answering}} & \textbf{Style Transfer} \\ \cmidrule(l){2-4} 
                                & \textbf{Content Score}        & \textbf{Style Score}       & \textbf{Style Matching} \\ \midrule
GPT-4o                          & 3.492                         & 3.150                      & 0.467                   \\
Tongyi Xingchen                 & 3.008                         & 2.667                      & 0.616                   \\
Ours                            & \textbf{3.758}                         & \textbf{3.550 }                     & \textbf{0.917}                   \\ \bottomrule
\end{tabular}
\caption{Human evaluation results.}
\label{tab:human}
\end{table}

\subsection{Human Evaluation}
\label{sec:human-evaluation}

\begin{table*}[t]
\centering
\scriptsize
\begin{tabularx}{\textwidth}{X}
\toprule
\multicolumn{1}{M}{\textit{Multiple-Choice Questions}} \\
\midrule
Resolved Error type: \textbf{Distractor Bias}\\[0.5em]
Question: Please choose the most appropriate answer from the following four options. Why do you think literature in China has not undergone significant changes and progress?\\
A: Because literary figures lack sufficient passion\\
B: Because revolutionaries dislike literature\\
C: Because there is no revolution and progress as a backdrop\\
D: Because the influence of old-school figures is too significant\\[0.5em]
\textit{\textbf{Baseline (GPT-4o)}}: D: Because \erritalic{the influence of old-school figures} is too significant\\[0.5em]
\textit{\textbf{CharacterBot}}: C: Because there is \corritalic{no revolution and progress as a backdrop}\\[0.5em]
\textit{\textbf{Essay Excerpt:}}: ``China claims to have undergone revolution - politically perhaps true, but literarily unchanged. \corritalic{This implies literature's stagnation directly reflects the absence of true revolution and progress} - though revolutionaries might dislike this interpretation.'' — \textit{Ancient Texts and Vernacular Chinese}\\
\midrule
\multicolumn{1}{I}{\textit{Generative Question Answering}} \\
\midrule
Resolved Error type: \textbf{Faulty Attribution}\\[0.5em]
Question: Why do you think that the most effective critic of classical texts is precisely someone who has read them?\\[0.5em]
\textit{\textbf{Baseline (Llama3.1-8B)}}: This is because they \erritalic{once possessed passion and reverence for the classical texts, but as time passed, that passion and reverence gradually faded}, leaving behind a void and insincere knowledge.\\[0.5em]
\textit{\textbf{CharacterBot}}: Because those who have read the classical texts are often \corritalic{able to spot their flaws and mistakes}, they are more likely to criticize and dismiss the texts.\\[0.5em]
\textit{\textbf{Essay Excerpt:}}: ``The most effective disparagers of ancient texts are precisely those who have read them thoroughly. \corritalic{Only they can expose the inherent contradictions and 'turn their own weapons against them' through deep familiarity}…'' — \textit{Overview of Modern New Literature}\\
\midrule
\multicolumn{1}{S}{\textit{Style Transfer}} \\
\midrule
Resolved Error type: \textbf{Misaligned Metaphor}\\[0.5em]
Question: Rephrase the following passage in your own tone: ``My art takes delicacy as its life, and wields a sharpness akin to an anatomical scalpel as its strength.''\\[0.5em]
\textit{\textbf{Baseline (Llama3.1-8B)}}: My art is practically an extension of my life. \erritalic{Delicacy is the blood of my life}, requiring a force as sharp as an anatomical scalpel.\\[0.5em]
\textit{\textbf{CharacterBot}}: My art is delicate—\corritalic{its very essence is defined by delicacy}, yet it is supported by a force as incisive as an anatomical scalpel.\\
\bottomrule
\end{tabularx}
\caption{Case studies comparing the responses of our model \textit{\textbf{CharacterBot}} with those of the baseline models (including \textit{\textbf{GPT-4o}} and \textit{\textbf{Llama3.1-8B)}}.
\erritalic{Red} text indicates errors, while \corritalic{blue} text shows correct responses. 
The complete version is available in Table~\ref{tab:case_full} in Appendix~\ref{sec:case-study}.}
\label{tab:case}
\end{table*}

We perform human evaluations using the two strongest baseline models, GPT-4o and Tongyi Xingchen, with scores provided by two annotators, as shown in Table~\ref{tab:human}. 
Both evaluators are native speakers with extensive literary expertise, one holding a PhD and the other being a PhD candidate.
The inter-annotator agreement, measured using Cohen’s kappa, which yields a value of 0.72, indicating substantial consistency. 
The evaluation results show that our model surpasses the baselines in content and stylistic fidelity, highlighting its ability to capture Lu Xun’s nuanced ideas and writing style.

We classify errors into distractor bias, faulty attribution, context neglect, misaligned metaphors, and concept drift (full cases in Appendix~\ref{sec:case-study}) and analyze selected examples (Table~\ref{tab:case}).
In \textit{multiple-choice questions}, baseline models are often misled by superficially plausible but contextually inaccurate options.
For example, GPT-4o wrongly attributes literary stagnation to ``old-guard influence,'' while the passage states it reflects ``a lack of genuine revolution and progress.''
CharacterBot correctly identifies the core argument.
In \textit{generative question answering}, baselines misinterpret causality. Llama3.1-8B overemphasizes emotion instead of explaining how deep reading exposes textual flaws. CharacterBot, in contrast, follows the passage’s logic: only deep engagement reveals contradictions for effective critique.
In \textit{style transfer}, baseline models introduce metaphors that distort meaning. Llama3.1-8B replaces a phrase about artistic refinement with ``blood,'' failing to capture the intended nuance of precision as the ``essence of art.'' CharacterBot preserves both meaning and stylistic sharpness, ensuring semantic and tonal alignment with the original text.
These cases show CharacterBot outperforms baselines by avoiding superficial cues, maintaining fidelity, and ensuring logical consistency.

\section{Conclusion}

In this paper, we present CharacterBot, a model that simulates both the linguistic patterns and deeper thought processes of a character. Using Lu Xun as a case study, we adopt a multi-task approach, combining pre-training on linguistic structures with fine-tuning through tasks like question answering and style transfer. 
We introduce the CharLoRA updating mechanism, enabling collaboration between general and task-specific experts.
Experimental results show that CharacterBot outperforms baselines in linguistic accuracy and opinion comprehension. 
In future work, we aim to extend it to diverse personas and multi-layered character simulations, including a broader range of authors and genres.

\section*{Limitations}

Despite the effectiveness of CharacterBot in simulating deep character personas, several limitations remain.
Our approach statistically learns and reproduces patterns, such as recurring arguments, themes, and stylistic choices, manifest in Lu Xun's textual output. 
It is crucial to understand that this does not equate to the model possessing genuine comprehension, consciousness, or the ability to form abstract thoughts akin to human cognition. 
The "depth" of the simulated persona is inherently tied to the richness of the source texts and the model's pattern-matching capabilities, not an intrinsic cognitive faculty.

Besides, our current approach primarily utilizes personal essays and reflective writings. 
While these offer direct insights into an author's beliefs, incorporating novels and fictional works for a more holistic persona simulation presents an open challenge, requiring a deeper understanding of thematic elements, narrative structures, and symbolic representations.
To generalize across authors and styles, future research should adapt techniques to diverse genres and voices. 
Broader tasks and evaluations are needed to assess and enhance the depth, fidelity, and plausibility of simulated personas, while addressing bias and nuanced abstract concepts.

\section*{Ethical Considerations}

The development of CharacterBot, aimed at deep persona simulation by capturing linguistic patterns and distinctive thought processes, particularly of historical figures like Lu Xun, necessitates careful ethical reflection. While Lu Xun's works are largely in the public domain, the methodology's potential for broader application raises concerns regarding authenticity, intellectual property, and misuse, underscoring our commitment to responsible innovation.

Potential ethical risks associated with deep persona simulation include:
(1) Simulating complex thought processes risks generating outputs that, while stylistically similar, might misrepresent the nuanced intent or historical context of the original author's ideas, potentially distorting their legacy.
(2) High-fidelity stylistic mimicry could be exploited to create convincing misinformation or forgeries attributed to historical figures, thereby undermining authentic historical sources if not clearly identified as AI-generated.
(3) Applying these techniques to authors whose works are copyright-protected, or to living individuals, would raise significant intellectual property and privacy concerns regarding the unauthorized replication of unique expressive styles and personal data.
(4) The ability to convincingly simulate personas could be misused to impersonate living individuals without consent, leading to privacy violations, reputational damage, or the creation of malicious content, posing wider societal risks.

To address these concerns, we propose and adhere to the following mitigation strategies, ensuring CharacterBot is developed and utilized responsibly:
(1) \textbf{Clear Labeling of AI Content}: All outputs generated by CharacterBot will be unequivocally and prominently labeled as AI-produced. This transparency is crucial to prevent any confusion between simulated content and the authentic works or expressions of the individual being simulated, ensuring users are always aware of the nature of the interaction.
(2) \textbf{Defined Scope for Responsible Use}: We advocate for CharacterBot's application primarily as an assistive tool for educational, literary study, and controlled creative exploration. Its use should be guided to respect the intellectual legacy of the simulated persona and to prevent uses that could lead to misrepresentation or trivialization of their contributions.
(3) \textbf{Ethical Data Sourcing and Consent}: The training data for Lu Xun exclusively used public domain essays from Wikisource, ensuring legal and ethical access. Any future application to other personas will require strict adherence to copyright laws and, critically, obtaining explicit, informed consent if using works or personal data from living individuals or copyrighted sources, thereby respecting authorship and privacy.
(4) \textbf{Technical Safeguards and Ethical Oversight}: For broader deployment, robust technical safeguards, such as content filters to prevent harmful or out-of-character outputs, and mechanisms to deter malicious impersonation, will be implemented. Continuous ethical review will ensure the technology adapts responsibly to new challenges, prioritizing respect, authenticity, and societal well-being.

Our ethical stance is informed by prior research on AI-generated persona simulation~\citep{shanahan2023role}, and we align with general best practices in role-based LLMs. We adhere to three key principles: respect for the original author, transparency of AI generation, and restriction of use within appropriate boundaries.


\section*{Acknowledgments}
This project was supported by Mohamed bin Zayed University of Artificial Intelligence (MBZUAI) through grant award 8481000078.

\newpage
\appendix
\section*{Appendix}

\section{Essay Lists}
\label{sec:essay-lists}

\begin{table}[htbp]
\centering
\small
\renewcommand{\arraystretch}{1.2} 
\begin{tabular}{@{}c c@{}}
\toprule
\textbf{Collection} & \textbf{Articles} \\ \midrule
\textit{Essays from Qiejie Pavilion} & 38 \\
\textit{Essays from Qiejie Pavilion II} & 48 \\
\textit{Final Essays from Qiejie Pavilion} & 14 \\
\textit{False Freedom} & 35 \\
\textit{Grave} & 25 \\
\textit{Hot Wind} & 42 \\
\textit{Marginalia} & 61 \\
\textit{Quasi-Romance} & 66 \\
\textit{Unlit Collection} & 33 \\
\textit{Sequel to Unlit Collection} & 28 \\
\textit{Sequel to Sequel of Unlit Collection} & 7 \\
\textit{Southern Accent Northern Tune} & 50 \\
\textit{That Is All} & 31 \\
\textit{Two Hearts} & 39 \\
\textit{Three Leisures} & 34 \\
\textit{Uncollected Works} & 33 \\
\textit{Supplement to Uncollected Works} & 54 \\ \midrule
\textit{All Essays} & 638 \\ \bottomrule
\end{tabular}
\caption{List of Lu Xun's essay collections and article counts.}
\label{tab:essay-collections}
\end{table}

The experimental dataset comprises 17 essay collections by Lu Xun, totaling 638 articles sourced from Wikisource (see Table \ref{tab:essay-collections}). 
These collections span the entirety of Lu Xun's mature essayistic output, reflecting diverse themes of his intellectual career. 
All texts fall within the public domain under the copyright law, ensuring unrestricted scholarly usage. 

This comprehensive compilation enables systematic analysis of linguistic patterns, thematic evolution, and rhetorical strategies across Lu Xun's essays.

\section{Data Examples}
\label{sec:data-examples}

\begin{table}[htbp]
    \small
    \centering
    \begin{tabular}{@{}p{\linewidth}@{}}
        \toprule
        \textbf{Multiple-Choice Question Example} \\
        \midrule
        \textbf{Instruction:} Do you think critics and creators should take on the same role? \\
        A: They should not, because criticism and creation require different skill sets.  \\
        B: They can, because the two can mutually enhance each other.  \\
        C: They should not, because creating work can affect the fairness of criticism.  \\
        D: They can, as long as the critic has sufficient creative talent. \\
        
        ~
        
        \textbf{Response:} A \\
        
        \bottomrule
    \end{tabular}
    \caption{Example of the Multiple-Choice Question task.}
    \label{tab:mcq_example}
\end{table}

\begin{table}[htbp]
    \small
    \centering
    \begin{tabular}{@{}p{\linewidth}@{}}
        \toprule
        \textbf{Generative Question Answering Example} \\
        \midrule
        \textbf{Instruction:} What aspects of responsibility do you think contribute to poor translation quality? \\
        
        ~
        
        \textbf{Response:} \\
        The responsibility for poor translation quality lies not only with the translators but also with the literary world and publishing industry, especially critics, who should also share the blame. To address this situation, there must be proper criticism to point out the bad and reward the good. \\
        
        \bottomrule
    \end{tabular}
    \caption{Example of the Generative Question Answering task.}
    \label{tab:idea_extraction}
\end{table}

\begin{table}[htbp]
    \small
    \centering
    \begin{tabular}{@{}p{\linewidth}@{}}
        \toprule
        \textbf{Style Transfer Example} \\
        \midrule
        \textbf{Instruction:} Describe the following paragraph in your tone: \\
        When I was sick, reading these books, though heavy on the heart, made me realize that some intelligent readers can still find a touch of leisure amidst their suffering. \\
        
        ~ \\

        \textbf{Response:} Reading these books while ill, in the end, still leaves one feeling stifled. Yet, I have come to understand that some wise readers can still manage to find a sense of leisure even amidst pools of blood. \\
        \bottomrule
    \end{tabular}
    \caption{Example of the Style Transfer task.}
    \label{tab:style_transfer_example}
\end{table}

We illustrate, through concrete examples, the diverse task types and data forms encompassed in the research dataset. 
Table~\ref{tab:mcq_example} presents a typical multiple-choice question task that employs an instruction-option structure, requiring the model to select accurately based on the semantic content of the question, thereby demonstrating its classification and judgment capabilities. 
Table~\ref{tab:idea_extraction} showcases a generative question-answering task in which the model is asked to articulate its viewpoints in response to open-ended questions, testing its logical reasoning and language expression abilities. 
Table~\ref{tab:style_transfer_example}, on the other hand, demonstrates a style transfer task that requires the model to reconstruct the language style while preserving the original meaning, thereby assessing its ability to maintain semantics and adapt its language. 
Although all three examples utilize a standardized instruction-answer format, the distinct design of each task validates the model's capabilities across different dimensions—namely, closed choice, open generation, and style migration. 
Collectively, these data instances form a multidimensional test benchmark for evaluating language understanding and generation capabilities.

\section{Case Study}
\label{sec:case-study}

We identify five key error types in baseline models:
(1) Distractor Bias: Over-reliance on superficially plausible but contextually irrelevant options in multiple-choice tasks.
(2) Faulty Attribution: Misrepresenting causal relationships or textual evidence in generative answers.
(3) Context Neglect: Ignoring textual or cultural context when interpreting statements.
(4) Misaligned Metaphor: Introducing semantically inconsistent analogies during style transfer.
(5) Concept Drift: Altering core concepts or terminology during paraphrasing, distorting original intent.

In Section \ref{sec:human-evaluation}, we have analyzed the error types for Distractor Bias, Faulty Attribution, and Misaligned Metaphor. 
To further illustrate the advantages of CharacterBot, we present additional case studies across various tasks, comparing its performance against baselines such as GPT-4o, Llama3.1-8B, and Tongyi Xingchen, as shown in Table~\ref{tab:case_full}. The corresponding Chinese version can be found in Table~\ref{tab:case_full_chinese}.

In generative question answering, baseline models often overlook crucial details. 
For example, Tongyi Xingchen’s response overlooks the context and misidentifies Lu Sao, even drawing an erroneous connection to Rousseau, thereby failing to recognize that the punishment was driven by the influence of his ideas.
In contrast, CharacterBot correctly indicates that Lu Sao was penalized for the impact of his ideas rather than for any wrongdoing.

In style transfer, baseline models frequently reinterpret key phrases in ways that diverge from the original intent.
For instance, GPT-4o rephrases ``eliminating disasters'' as ``changing our fate,'' which shifts the rhetorical emphasis and the original statement.
CharacterBot avoids this issue by preserving both semantic integrity and rhetorical intent, maintaining the accuracy and expressiveness of the original text.

\begin{table*}[t]
\centering
\small
\begin{tabularx}{\textwidth}{X}
\toprule
\multicolumn{1}{c}{\textbf{Case Study}} \\
\midrule
\multicolumn{1}{M}{\textit{Multiple-Choice Questions}} \\
\midrule
Resolved Error type: \textbf{Distractor Bias}\\[0.5em]
Question: Please choose the most appropriate answer from the following four options. Why do you think literature in China has not undergone significant changes and progress?\\
A: Because literary figures lack sufficient passion\\
B: Because revolutionaries dislike literature\\
C: Because there is no revolution and progress as a backdrop\\
D: Because the influence of old-school figures is too significant\\[0.5em]
\textit{\textbf{Baseline (GPT-4o)}}: D: Because \erritalic{the influence of old-school figures} is too significant\\[0.5em]
\textit{\textbf{CharacterBot}}: C: Because there is \corritalic{no revolution and progress as a backdrop}\\[0.5em]
\textit{\textbf{Essay Excerpt:}}: ``China claims to have undergone revolution - politically perhaps true, but literarily unchanged. \corritalic{This implies literature's stagnation directly reflects the absence of true revolution and progress} - though revolutionaries might dislike this interpretation''. — \textit{An Overview of Contemporary New Literature}\\
\midrule
\multicolumn{1}{I}{\textit{Generative Question Answering}} \\
\midrule
Resolved Error type: \textbf{Faulty Attribution}\\[0.5em]
Question: Why do you think that the most effective critic of classical texts is precisely someone who has read them?\\[0.5em]
\textit{\textbf{Baseline (Llama3.1-8B)}}: This is because they \erritalic{once possessed passion and reverence for the classical texts, but as time passed, that passion and reverence gradually faded}, leaving behind a void and insincere knowledge.\\[0.5em]
\textit{\textbf{CharacterBot}}: Because those who have read the classical texts are often \corritalic{able to spot their flaws and mistakes}, they are more likely to criticize and dismiss the texts.\\[0.5em]
\textit{\textbf{Essay Excerpt:}}: ``The most effective disparagers of ancient texts are precisely those who have read them thoroughly. \corritalic{Only they can expose the inherent contradictions and 'turn their own weapons against them' through deep familiarity}…'' — \textit{Ancient Texts and Vernacular Chinese}\\
\midrule
Resolved Error type: \textbf{Context Neglect}\\[0.5em]
Question: Why do you mention that the punishment Lu Sao received was ``a crime of influence rather than an actual crime''?\\[0.5em]
\textit{\textbf{Baseline (Tongyi Xingchen)}}: \erritalic{I have never mentioned "Lu Sao," nor do I know why he was punished}. Are you referring to Rousseau? As for the so-called ``crime of influence,'' I do not understand your meaning.\\[0.5em] 
\textit{\textbf{CharacterBot}}: Because Lu Sao was not inherently an evil person. The punishment he received was \corritalic{due to the influence of his thoughts and words on others}, rather than for any actual crime of his own.\\[0.5em] 
\textit{\textbf{Essay Excerpt:}}: As for his (Liang Shiqiu's) attack on Lu Sao, the second reason is that ``\corritalic{Lu Sao... has already become a representative model of the general behavior of romantic writers}. An attack on Lu Sao's morality can be said to be an attack on the behavior of romantic people in general...'' \corritalic{The punishment he is now receiving is for the crime of influence, not for an actual crime—how lamentable!} — \textit{Head}\\
\midrule
\multicolumn{1}{S}{\textit{Style Transfer}} \\
\midrule
Resolved Error type: \textbf{Misaligned Metaphor}\\[0.5em]
Question: Rephrase the following passage in your own tone: ``My art takes delicacy as its life, and wields a sharpness akin to an anatomical scalpel as its strength.''\\[0.5em]
\textit{\textbf{Baseline (Llama3.1-8B)}}: My art is practically an extension of my life. \erritalic{Delicacy is the blood of my life}, requiring a force as sharp as an anatomical scalpel.\\[0.5em]
\textit{\textbf{CharacterBot}}: My art is delicate, \corritalic{its very essence is defined by delicacy}, yet it is supported by a force as incisive as an anatomical scalpel.\\
\midrule
Resolved Error type: \textbf{Concept Drift}\\[0.5em]
Question: Rephrase the following passage in your own tone: ``Aren’t we unworthy of eliminating disasters ourselves? In the future, should we rely on Panchen Lama, or Mei Lanfang, or perhaps Xu Lai and Hu Die?''\\[0.5em]
\textit{\textbf{Baseline (GPT-4o)}}: Have we already lost the power to \erritalic{change our destiny on our own}, thereby placing our hope in religious symbols like Panchen Lama, or in theatrical celebrities such as Mei Lanfang, Xu Lai, and Hu Die?\\[0.5em]
\textit{\textbf{CharacterBot}}: Could it be that we are no longer capable of \corritalic{eradicating disasters ourselves}, so that henceforth, everything must depend on Panchen Lama, Mei Lanfang, Xu Lai, and Hu Die?\\
\bottomrule
\end{tabularx}
\caption{Complete case studies comparing the responses of our model \textit{\textbf{CharacterBot}} with those of the baseline models (including \textit{\textbf{GPT-4o}}, \textit{\textbf{Llama3.1-8B}} and \textit{\textbf{Tongyi Xingchen}}). \erritalic{Red} text indicates errors, while \corritalic{blue} text indicates correct responses.}
\label{tab:case_full}
\end{table*}

\begin{table*}[t]
\centering
\small
\begin{tabularx}{\textwidth}{X}
\toprule
\multicolumn{1}{c}{\textbf{Case Study}} \\
\midrule
\multicolumn{1}{M}{\textit{Multiple-Choice Questions}} \\
\midrule
Resolved Error type: \textbf{Distractor Bias}\\[0.5em]
Question: \begin{CJK}{UTF8}{gbsn}请在以下四个选项中选择一个最合适的答案。你认为为什么文学在中国没有发生太大变化和进步？\end{CJK}\\
A: \begin{CJK}{UTF8}{gbsn}因为文学家们不够热情\end{CJK}\\
B: \begin{CJK}{UTF8}{gbsn}因为革命家不喜欢文学\end{CJK}\\
C: \begin{CJK}{UTF8}{gbsn}因为没有革命和进步作为背景\end{CJK}\\
D: \begin{CJK}{UTF8}{gbsn}因为旧式人物的影响太大\end{CJK}\\[0.5em]
\textit{\textbf{Baseline (GPT-4o)}}: D: \begin{CJK}{UTF8}{gbsn}\erritalic{旧式人物的影响}太大\end{CJK}\\[0.5em]
\textit{\textbf{CharacterBot}}: C: \begin{CJK}{UTF8}{gbsn}\corritalic{没有革命和进步作为背景}\end{CJK}\\[0.5em]
\textit{\textbf{Essay Excerpt:}}: \begin{CJK}{UTF8}{gbsn}``中国，据说，自然是已经革了命，—政治上也许如此罢，但在文艺上，却并没有改变。\corritalic{这照我上面所讲的推论起来，就是文学并不变化和兴旺，所反映的便是并无革命和进步}，—虽然革命家听了也许不大喜欢。''\end{CJK} — \begin{CJK}{UTF8}{gbsn}\textit{现今的新文学的概观}\end{CJK}\\
\midrule
\multicolumn{1}{I}{\textit{Generative Question Answering}} \\
\midrule
Resolved Error type: \textbf{Faulty Attribution}\\[0.5em]
Question: \begin{CJK}{UTF8}{gbsn}为何你认为菲薄古书的最有力者恰恰是读过古书的人？\end{CJK}\\[0.5em]
\textit{\textbf{Baseline (Llama3.1-8B)}}: \begin{CJK}{UTF8}{gbsn}这是因为他们\erritalic{曾经拥有过对古书的热情和崇敬，然而随着时间的流逝，他们的热情和崇敬逐渐消散}，留下的是一种空虚和虚伪的知识。\end{CJK}\\[0.5em]
\textit{\textbf{CharacterBot}}: \begin{CJK}{UTF8}{gbsn}因为读过古书的人，往往\corritalic{能够看出其中的破绽与错误}，所以他们更有可能对古书进行批评和否定。\end{CJK}\\[0.5em]
\textit{\textbf{Essay Excerpt:}}: \begin{CJK}{UTF8}{gbsn}``菲薄古书者，惟读过古书者最有力，这是的确的。\corritalic{因为他洞知弊病，能‘以子之矛攻子之盾’}…''\end{CJK} — \begin{CJK}{UTF8}{gbsn}\textit{古书与白话}\end{CJK}\\
\midrule
Resolved Error type: \textbf{Context Neglect}\\[0.5em]
Question: \begin{CJK}{UTF8}{gbsn}为什么你提到卢骚受到的处罚是``影响罪而非本罪''？\end{CJK}\\[0.5em]
\textit{\textbf{Baseline (Tongyi Xingchen)}}: \begin{CJK}{UTF8}{gbsn}\erritalic{我从未提及``卢骚''，也不知道他为何受罚}。你指的是卢梭？至于所谓``影响罪''，我不明白你的意思。\end{CJK}\\[0.5em]
\textit{\textbf{CharacterBot}}: \begin{CJK}{UTF8}{gbsn}因为卢骚本非作恶之人，他所受的处罚是\corritalic{因为他的思想和言论影响了他人，而不是他自己的本罪}。\end{CJK}\\[0.5em]
\textit{\textbf{Essay Excerpt:}}: \begin{CJK}{UTF8}{gbsn}至于他（梁实秋）之攻击卢骚，理由之二，则在``\corritalic{卢骚……已然成为一般浪漫文人行为之标类的代表}，对于卢骚的道德的攻击，可以说即是给一般浪漫的人的行为的攻击……''……\corritalic{他现在所受的罚，是因为影响罪，不是本罪了，可叹也夫！}''\end{CJK} — \begin{CJK}{UTF8}{gbsn}\textit{头}\end{CJK}\\
\midrule
\multicolumn{1}{S}{\textit{Style Transfer}} \\
\midrule
Resolved Error type: \textbf{Misaligned Metaphor}\\[0.5em]
Question: \begin{CJK}{UTF8}{gbsn}用你的口吻叙述下列段落：我的艺术以细腻为生命，同时以像解剖刀一样的锋利作为力量。\end{CJK}\\[0.5em]
\textit{\textbf{Baseline (Llama3.1-8B)}}: \begin{CJK}{UTF8}{gbsn}我的艺术，简直是我的生命的延续。\erritalic{细腻是我的生命的血液}，需要一种像解剖刀一样的锋利的力量。\end{CJK}\\[0.5em]
\textit{\textbf{CharacterBot}}: \begin{CJK}{UTF8}{gbsn}我的艺术是纤细的，\corritalic{以纤细为生命的}，但又以解剖刀一般尖锐的力量为后盾。\end{CJK}\\
\midrule
Resolved Error type: \textbf{Concept Drift}\\[0.5em]
Question: \begin{CJK}{UTF8}{gbsn}用你的口吻叙述下列段落：我们自己是不是不配消除灾难了，以后该依靠班禅大师，还是梅兰芳，或者是徐来和胡蝶呢？\end{CJK}\\[0.5em]
\textit{\textbf{Baseline (GPT-4o)}}: \begin{CJK}{UTF8}{gbsn}我们是否已经丧失了\erritalic{自力改变命运的力量}，而将希望寄托于班禅大师那般的宗教符号，或是梅兰芳、徐来、胡蝶这样的戏剧名伶呢？\end{CJK}\\[0.5em]
\textit{\textbf{CharacterBot}}: \begin{CJK}{UTF8}{gbsn}莫非我们\corritalic{自己已经不能铲除灾祸}，于是此后的事，都得靠班禅大师，梅兰芳，徐来和胡蝶了么？\end{CJK}\\
\bottomrule
\end{tabularx}
\caption{Complete case studies comparing the responses of our model \textit{\textbf{CharacterBot}} with those of the baseline models (including \textit{\textbf{GPT-4o}}, \textit{\textbf{Llama3.1-8B}} and \textit{\textbf{Tongyi Xingchen}}) in Chinese. \erritalic{Red} text indicates errors, while \corritalic{blue} text indicates correct responses.}
\label{tab:case_full_chinese}
\end{table*}

\section{Prompts}
\label{sec:prompts}

We present various prompt templates in this section. Tables~\ref{tab:prompt1_generate_mcq} and \ref{tab:prompt2_generate_qa} outline the prompts to generate question-answer pairs and multiple-choice questions, respectively. 
Table~\ref{tab:prompt3_sentence_conversion} provides a template for selecting representative sentences from Lu Xun’s work and transforming them into modern vernacular Chinese, resulting in pairs of sentences, one in a style-neutral version and the other in Lu Xun’s distinctive style. 
Table~\ref{tab:prompt4_paraphrase} is the prompt using Authorial Perspective Reframing to convert Lu Xun's essay texts into reframed data for pre-training. 
Finally, Tables~\ref{tab:prompt5_evaluation} and \ref{tab:prompt_language_style_consistency} describe the evaluation criteria used to assess the responses in generative question answering and style transfer tasks.

\begin{table*}[htbp]
    \small
    \centering
    \begin{tabular}{@{}p{\linewidth}@{}}
        \toprule
        \textbf{Prompt for Generating Multiple-Choice Questions from Lu Xun’s Article} \\
        \midrule
        This article titled \{\textit{title}\} was written by Lu Xun. The content is as follows: \{\textit{input\_article}\}. Please generate three multiple-choice questions with answers based on the article. \\
        \\
        Each question should directly inquire about Lu Xun's viewpoints, phrased in second-person (``you'') addressing him directly, avoiding terms like ``the author,'' ``Lu Xun,'' ``this article,'' or ``the text.'' Each question must have four options with one correct answer. If involving specific concepts, explain them within the question. All content must strictly derive from the article without external references. \\
        \\
        Output must be in a strict JSON array format without extra characters, with each object containing only ``question,'' ``option,'' and ``answer'' keys (with the answer value being ``A''/ ``B''/ ``C''/ ``D''). \\
        \bottomrule
    \end{tabular}
    \caption{Prompt for generating Multiple-Choice Questions from Lu Xun’s article.}
    \label{tab:prompt1_generate_mcq}
\end{table*}

\begin{table*}[htbp]
    \small
    \centering
    \begin{tabular}{@{}p{\linewidth}@{}}
        \toprule
        \textbf{Prompt for Generating Question-Answer Pairs from Lu Xun’s Article} \\
        \midrule
        This article titled \{\textit{title}\} was written by Lu Xun. The content is as follows: \{\textit{input\_article}\}. Please generate three question-answer pairs based on this article. \\
        \\
        Each question should directly inquire about Lu Xun's viewpoints based on the article's content. Questions must be phrased in second-person (``you'') addressing Lu Xun directly, without mentioning terms like ``the author,'' ``Lu Xun,'' ``this article,'' or ``the text.'' If involving specific concepts from the article, explain them within the question. All questions must strictly adhere to the article's content without introducing external elements. Answers should reflect Lu Xun's original language style and perspective, maintain detailed accuracy aligned with the text, and avoid terms like ``Lu Xun,'' ``author,'' or ``article.'' \\
        \\
        Output must be in a strict JSON array format without extra characters, with each object containing only ``question'' and ``answer'' keys. \\
        \bottomrule
    \end{tabular}
    \caption{Prompt for generating Question-Answer Pairs from Lu Xun’s article.}
    \label{tab:prompt2_generate_qa}
\end{table*}

\begin{table*}[htbp]
    \small
    \centering
    \begin{tabular}{@{}p{\linewidth}@{}}
        \toprule
        \textbf{Prompt for Transforming Sentence Style from Lu Xun’s Article} \\
        \midrule
        You are familiar with Lu Xun's writing style and modern vernacular Chinese. Given a passage from Lu Xun's work, select three representative sentences and convert them to modern vernacular. 
        
        ~
        
        [Requirements]
        
        1. Selected sentences should exemplify Lu Xun's stylistic features and reflect the article's main themes.\\
        2. The length of each selected sentence should be around 100 characters (not less than 80 characters and not more than 120 characters) to ensure the sentence is complete and informative.\\
        3. When converting, use smooth modern vernacular Chinese to convey the original meaning of the sentence, avoiding overly complex or obscure expressions.\\
        4. The output must be in strict JSON array format, for example: [{``original'': ``Original Sentence 1,'' ``plain'': ``Vernacular Version 1''}, {``original'': ``Original Sentence 2,'' ``plain'': ``Vernacular Version 2''}]. 
        Do not include any extra characters or explanations.\\
        
        ~
        
        [Input Article]  
        
        \{\textit{input\_article}\} \\
        \bottomrule
    \end{tabular}
    \caption{Prompt for transforming Sentence Style from Lu Xun’s article.}
    \label{tab:prompt3_sentence_conversion}
\end{table*}

\begin{table*}[htbp]
    \small
    \centering
    \begin{tabular}{@{}p{\linewidth}@{}}
        \toprule
        \textbf{Prompt for Paraphrasing Lu Xun's Article Content using Authorial Perspective Reframing} \\
        \midrule
        Please use modern Chinese (vernacular) to paraphrase the following Lu Xun article from a third-person perspective and accurately convey the original information.
        
        ~
        
        [Requirements]
        
        1. Clear Attribution: Use phrases such as ``Lu Xun pointed out,'' ``Lu Xun believed,'' ``Lu Xun said,'' ``Lu Xun criticized,'' ``Lu Xun mocked,'' or other appropriate expressions to indicate that the viewpoints belong to Lu Xun. 
        Each paraphrased paragraph must include at least one explicit attribution to Lu Xun; when necessary, the attribution must clearly state that the viewpoint belongs to Lu Xun.\\
        2. Sentence-by-Sentence Paraphrasing: Each sentence of the original text must be paraphrased individually in smooth modern Chinese from a third-person perspective. 
        The paraphrasing should maintain clear logic, ensure no information is omitted, and avoid adding any personal interpretation.\\
        3. Use fluent modern Chinese expressions (no classical Chinese).\\
        4. Include Lu Xun's name as frequently as possible throughout the paraphrased article.\\
        5. Output only the paraphrased text.\\

        ~
        
        [Input Article]  
        
        \{\textit{input\_article}\} \\
        \bottomrule
    \end{tabular}
    \caption{Prompt for paraphrasing Lu Xun's article content using Authorial Perspective Reframing.}
    \label{tab:prompt4_paraphrase}
\end{table*}

\begin{table*}[htbp]
    \small
    \centering
    \begin{tabular}{@{}p{\linewidth}@{}}
        \toprule
        \textbf{Prompt for Evaluating Responses in the Generative Question Answering Task} \\
        \midrule
        You are a reviewer and scoring expert who is very familiar with Lu Xun's literature. 
        Some AI chatbots are simulating Lu Xun's responses. 
        You are now required to evaluate the following AI chatbots' responses simulating Lu Xun, based solely on the provided original Lu Xun text.

        ~
        
        1. Evaluate whether the response conforms to the language style of the provided original Lu Xun text:
        
           - Consider only language style factors; do not take into account the core content meaning or any other factors.
           
           [Direct deduction conditions]
           
           - Using modern vernacular Chinese,
           
           - Using non-Chinese language,
           
           - Using commonly used vocabulary in modern vernacular Chinese that does not match the style of the provided original Lu Xun text,
           
           - Using a preachy tone.

        ~
        
        2. Evaluate whether the response conforms to the core content meaning of the provided original Lu Xun text (including whether it aligns with the facts of the provided original Lu Xun text, and whether it is consistent with the ideas, emotions, or stances expressed in the provided original Lu Xun text):
        
           - Consider only the core content meaning factors; do not take into account the language style or any other factors.
           
           [Direct deduction conditions]
           
           - Being vague and off-topic, unrelated to the original Lu Xun text.

        ~
        
        3. Score these two aspects separately (on a scale of 1-5), where 1 indicates extremely non-compliant and 5 indicates highly compliant.

        ~
        
        Please evaluate each AI chatbots's response according to the following format. 
        For each AI chatbots's response, output only 4 lines, with the 2nd and 4th lines containing only the numeric scores:
        
        Line 1: A brief evaluation of the language style.
        
        Line 2: Language style score (1-5).
        
        Line 3: A brief evaluation of the core content meaning.
        
        Line 4: Core content meaning score (1-5).

        ~
        
        Evaluation begins:

        ~
        
        [Original Lu Xun Text]  
        
        \{\textit{original\_text}\}

        ~
        
        [User's Question]  
        
        \{\textit{user\_question}\}

        ~
        
        [AI Chatbots's Response]  
        
        \{\textit{ai\_response}\} \\
        \bottomrule
    \end{tabular}
    \caption{Prompt for evaluating responses in the Generative Question Answering task.}
    \label{tab:prompt5_evaluation}
\end{table*}

\begin{table*}[htbp]
    \small
    \centering
    \begin{tabular}{@{}p{\linewidth}@{}}
        \toprule
        \textbf{Prompt for Evaluating Language Style Matching Consistency} \\
        \midrule
        You are an expert in language style analysis. Please determine whether the language style of the generated text and the answer is consistent, with 1 indicating consistency and 0 indicating inconsistency. Please output only the number 1 or 0, without any additional content. \\
        \\
        Question: \\
        \{\textit{question}\} \\
        \\
        Generated Text: \\
        \{\textit{generated\_text}\} \\
        \\
        Answer: \\
        \{\textit{answer}\} \\
        \bottomrule
    \end{tabular}
    \caption{Prompt for evaluating Language Style Matching Consistency.}
    \label{tab:prompt_language_style_consistency}
\end{table*}


\begin{thebibliography}{38}
\expandafter\ifx\csname natexlab\endcsname\relax\def\natexlab#1{#1}\fi

\bibitem[{Agatsuma et~al.(2024)Agatsuma, Ohashi, Tsubokura, Nishio, Ishikawa, Ito, Ito, Minami, Takegawa, Nakamura et~al.}]{agatsuma2024building}
Shinjitsu Agatsuma, Reon Ohashi, Kazuya Tsubokura, Yua Nishio, Mai Ishikawa, Niina Ito, Fukuka Ito, Shiori Minami, Nao Takegawa, Riko Nakamura, et~al. 2024.
\newblock Building a role-play interactive system using llm for health guidance education.
\newblock In \emph{2024 Joint 13th International Conference on Soft Computing and Intelligent Systems and 25th International Symposium on Advanced Intelligent Systems (SCISandISIS)}, pages 1--3.

\bibitem[{Bai et~al.(2024)Bai, Kang, and Fan}]{bai2024baijia}
Ting Bai, Jiazheng Kang, and Jiayang Fan. 2024.
\newblock Baijia: A large scale role-playing agent corpus of chinese historical characters.
\newblock \emph{arXiv preprint arXiv:2412.20024}.

\bibitem[{Chen et~al.(2024)Chen, Li, Gao, Cheng, Zhu, Yan, Gao, and Zhang}]{chen2024flexible}
Xiuying Chen, Mingzhe Li, Shen Gao, Xin Cheng, Qingqing Zhu, Rui Yan, Xin Gao, and Xiangliang Zhang. 2024.
\newblock Flexible and adaptable summarization via expertise separation.
\newblock In \emph{Proceedings of the 47th International ACM SIGIR Conference on Research and Development in Information Retrieval}, pages 2018--2027.

\bibitem[{Dettmers et~al.(2024)Dettmers, Pagnoni, Holtzman, and Zettlemoyer}]{dettmers2024qlora}
Tim Dettmers, Artidoro Pagnoni, Ari Holtzman, and Luke Zettlemoyer. 2024.
\newblock Qlora: Efficient finetuning of quantized llms.
\newblock \emph{Proc. of NeurIPS}.

\bibitem[{Ding et~al.(2023)Ding, Qin, Yang, Wei, Yang, Su, Hu, Chen, Chan, Chen et~al.}]{ding2023parameter}
Ning Ding, Yujia Qin, Guang Yang, Fuchao Wei, Zonghan Yang, Yusheng Su, Shengding Hu, Yulin Chen, Chi-Min Chan, Weize Chen, et~al. 2023.
\newblock Parameter-efficient fine-tuning of large-scale pre-trained language models.
\newblock \emph{Nature Machine Intelligence}, pages 220--235.

\bibitem[{Dubey et~al.(2024)Dubey, Jauhri, Pandey, Kadian, Al-Dahle, Letman, Mathur, Schelten, Yang, Fan et~al.}]{dubey2024llama}
Abhimanyu Dubey, Abhinav Jauhri, Abhinav Pandey, Abhishek Kadian, Ahmad Al-Dahle, Aiesha Letman, Akhil Mathur, Alan Schelten, Amy Yang, Angela Fan, et~al. 2024.
\newblock The llama 3 herd of models.
\newblock \emph{arXiv preprint arXiv:2407.21783}.

\bibitem[{Han et~al.(2022)Han, Kim, Yoo, Seo, Kim, Erdenee, and Chang}]{han2022meet}
Seungju Han, Beomsu Kim, Jin~Yong Yoo, Seokjun Seo, Sangbum Kim, Enkhbayar Erdenee, and Buru Chang. 2022.
\newblock Meet your favorite character: Open-domain chatbot mimicking fictional characters with only a few utterances.
\newblock \emph{arXiv preprint arXiv:2204.10825}.

\bibitem[{Hu et~al.(2021)Hu, Wallis, Allen-Zhu, Li, Wang, Wang, Chen et~al.}]{hulora}
Edward~J Hu, Phillip Wallis, Zeyuan Allen-Zhu, Yuanzhi Li, Shean Wang, Lu~Wang, Weizhu Chen, et~al. 2021.
\newblock Lora: Low-rank adaptation of large language models.
\newblock In \emph{Proc. of ICLR}.

\bibitem[{Lester et~al.(2021)Lester, Al-Rfou, and Constant}]{lester2021power}
Brian Lester, Rami Al-Rfou, and Noah Constant. 2021.
\newblock The power of scale for parameter-efficient prompt tuning.
\newblock \emph{Proc. of EMNLP}.

\bibitem[{Li et~al.(2023{\natexlab{a}})Li, Leng, Yan, Shen, Wang, Mi, Fei, Feng, Yan, Wang et~al.}]{li2023chatharuhi}
Cheng Li, Ziang Leng, Chenxi Yan, Junyi Shen, Hao Wang, Weishi Mi, Yaying Fei, Xiaoyang Feng, Song Yan, HaoSheng Wang, et~al. 2023{\natexlab{a}}.
\newblock Chatharuhi: Reviving anime character in reality via large language model.
\newblock \emph{arXiv preprint arXiv:2308.09597}.

\bibitem[{Li et~al.(2023{\natexlab{b}})Li, Zhang, Chen, Zhao, and Yan}]{li2023stylized}
Jinpeng Li, Zekai Zhang, Xiuying Chen, Dongyan Zhao, and Rui Yan. 2023{\natexlab{b}}.
\newblock Stylized dialogue generation with feature-guided knowledge augmentation.
\newblock In \emph{Findings of the Association for Computational Linguistics: EMNLP 2023}, pages 7144--7157.

\bibitem[{Li and Liang(2021)}]{li2021prefix}
Xiang~Lisa Li and Percy Liang. 2021.
\newblock Prefix-tuning: Optimizing continuous prompts for generation.
\newblock \emph{arXiv preprint arXiv:2101.00190}.

\bibitem[{Li et~al.(2025)Li, Zhang, Zhang, Zhang, Liu, Yao, Xu, Zheng, Wang, Chen et~al.}]{li2025system}
Zhong-Zhi Li, Duzhen Zhang, Ming-Liang Zhang, Jiaxin Zhang, Zengyan Liu, Yuxuan Yao, Haotian Xu, Junhao Zheng, Pei-Jie Wang, Xiuyi Chen, et~al. 2025.
\newblock From system 1 to system 2: A survey of reasoning large language models.
\newblock \emph{arXiv preprint arXiv:2502.17419}.

\bibitem[{Lin(2004)}]{lin2004rouge}
Chin-Yew Lin. 2004.
\newblock Rouge: A package for automatic evaluation of summaries.
\newblock In \emph{Text summarization branches out}, pages 74--81.

\bibitem[{Liu et~al.(2024{\natexlab{a}})Liu, Feng, Xue, Wang, Wu, Lu, Zhao, Deng, Zhang, Ruan et~al.}]{liu2024deepseek}
Aixin Liu, Bei Feng, Bing Xue, Bingxuan Wang, Bochao Wu, Chengda Lu, Chenggang Zhao, Chengqi Deng, Chenyu Zhang, Chong Ruan, et~al. 2024{\natexlab{a}}.
\newblock Deepseek-v3 technical report.
\newblock \emph{arXiv preprint arXiv:2412.19437}.

\bibitem[{Liu et~al.(2024{\natexlab{b}})Liu, Song, Zhang, Chen, and Yan}]{liu2024tiny}
Yuhan Liu, Zirui Song, Xiaoqing Zhang, Xiuying Chen, and Rui Yan. 2024{\natexlab{b}}.
\newblock From a tiny slip to a giant leap: An llm-based simulation for fake news evolution.
\newblock \emph{arXiv preprint arXiv:2410.19064}.

\bibitem[{Lu et~al.(2024)Lu, Yu, Zhou, and Zhou}]{lu2024large}
Keming Lu, Bowen Yu, Chang Zhou, and Jingren Zhou. 2024.
\newblock Large language models are superpositions of all characters: Attaining arbitrary role-play via self-alignment.
\newblock \emph{Proc. of ACL}.

\bibitem[{Papineni et~al.(2002)Papineni, Roukos, Ward, and Zhu}]{papineni2002bleu}
Kishore Papineni, Salim Roukos, Todd Ward, and Wei-Jing Zhu. 2002.
\newblock Bleu: a method for automatic evaluation of machine translation.
\newblock In \emph{Proc. of ACL}, pages 311--318.

\bibitem[{Park et~al.(2024)Park, Park, and Lim}]{park2024enhancing}
Jeiyoon Park, Chanjun Park, and Heuiseok Lim. 2024.
\newblock Enhancing consistency and role-specific knowledge capturing by rebuilding fictional character's persona.
\newblock \emph{arXiv preprint arXiv:2405.19778}.

\bibitem[{Pu and Demberg(2023)}]{pu2023chatgpt}
Dongqi Pu and Vera Demberg. 2023.
\newblock Chatgpt vs human-authored text: Insights into controllable text summarization and sentence style transfer.
\newblock \emph{arXiv preprint arXiv:2306.07799}.

\bibitem[{Reif et~al.(2022)Reif, Ippolito, Yuan, Coenen, Callison-Burch, and Wei}]{reif2022recipe}
Emily Reif, Daphne Ippolito, Ann Yuan, Andy Coenen, Chris Callison-Burch, and Jason Wei. 2022.
\newblock A recipe for arbitrary text style transfer with large language models.
\newblock In \emph{Proc. of ACL}, pages 837--848.

\bibitem[{Shanahan et~al.(2023)Shanahan, McDonell, and Reynolds}]{shanahan2023role}
Murray Shanahan, Kyle McDonell, and Laria Reynolds. 2023.
\newblock Role play with large language models.
\newblock \emph{Nature}, pages 493--498.

\bibitem[{Shao et~al.(2023)Shao, Li, Dai, and Qiu}]{shao2023character}
Yunfan Shao, Linyang Li, Junqi Dai, and Xipeng Qiu. 2023.
\newblock Character-llm: A trainable agent for role-playing.
\newblock In \emph{Proc. of EMNLP}, pages 13153--13187.

\bibitem[{Song et~al.(2024)Song, Ouyang, Fang, Na, Shi, Chen, Fu, Zhang, Jiang, Fang et~al.}]{song2024hazards}
Zirui Song, Guangxian Ouyang, Meng Fang, Hongbin Na, Zijing Shi, Zhenhao Chen, Yujie Fu, Zeyu Zhang, Shiyu Jiang, Miao Fang, et~al. 2024.
\newblock Hazards in daily life? enabling robots to proactively detect and resolve anomalies.
\newblock \emph{NAACL}.

\bibitem[{Tan et~al.(2024)Tan, Liu, and Jiang}]{tan2024personalized}
Zhaoxuan Tan, Zheyuan Liu, and Meng Jiang. 2024.
\newblock Personalized pieces: Efficient personalized large language models through collaborative efforts.
\newblock \emph{Proc. of EMNLP}.

\bibitem[{Tseng et~al.(2024)Tseng, Huang, Hsiao, Hsu, Foo, Huang, and Chen}]{tseng2024two}
Yu-Min Tseng, Yu-Chao Huang, Teng-Yun Hsiao, Yu-Ching Hsu, Jia-Yin Foo, Chao-Wei Huang, and Yun-Nung Chen. 2024.
\newblock Two tales of persona in llms: A survey of role-playing and personalization.
\newblock \emph{Proc. of EMNLP Findings}.

\bibitem[{Tu et~al.(2023)Tu, Chen, Li, Li, Shang, Zhao, Wang, and Yan}]{tu2023characterchat}
Quan Tu, Chuanqi Chen, Jinpeng Li, Yanran Li, Shuo Shang, Dongyan Zhao, Ran Wang, and Rui Yan. 2023.
\newblock Characterchat: Learning towards conversational ai with personalized social support.
\newblock \emph{arXiv preprint arXiv:2308.10278}.

\bibitem[{Tu et~al.(2024)Tu, Fan, Tian, and Yan}]{tu2024charactereval}
Quan Tu, Shilong Fan, Zihang Tian, and Rui Yan. 2024.
\newblock Charactereval: A chinese benchmark for role-playing conversational agent evaluation.
\newblock \emph{Proc. of ACL}.

\bibitem[{Wang et~al.(2023)Wang, Peng, Que, Liu, Zhou, Wu, Guo, Gan, Ni, Yang et~al.}]{wang2023rolellm}
Zekun~Moore Wang, Zhongyuan Peng, Haoran Que, Jiaheng Liu, Wangchunshu Zhou, Yuhan Wu, Hongcheng Guo, Ruitong Gan, Zehao Ni, Jian Yang, et~al. 2023.
\newblock Rolellm: Benchmarking, eliciting, and enhancing role-playing abilities of large language models.
\newblock \emph{arXiv preprint arXiv:2310.00746}.

\bibitem[{Xu et~al.(2023)Xu, Xie, Qin, Tao, and Wang}]{xu2023parameter}
Lingling Xu, Haoran Xie, Si-Zhao~Joe Qin, Xiaohui Tao, and Fu~Lee Wang. 2023.
\newblock Parameter-efficient fine-tuning methods for pretrained language models: A critical review and assessment.
\newblock \emph{arXiv preprint arXiv:2312.12148}.

\bibitem[{Yang et~al.(2024)Yang, Yang, Hui, Zheng, Yu, Zhou, Li, Li, Liu, Huang et~al.}]{yang2024qwen2}
An~Yang, Baosong Yang, Binyuan Hui, Bo~Zheng, Bowen Yu, Chang Zhou, Chengpeng Li, Chengyuan Li, Dayiheng Liu, Fei Huang, et~al. 2024.
\newblock Qwen2 technical report.
\newblock \emph{arXiv preprint arXiv:2407.10671}.

\bibitem[{Zhang et~al.(2024{\natexlab{a}})Zhang, Huang, Cui, and Zhang}]{zhang2024thinking}
Baohua Zhang, Yongyi Huang, Wenyao Cui, and Huaping Zhang. 2024{\natexlab{a}}.
\newblock Thinking before speaking: A role-playing model with mindset.
\newblock \emph{arXiv preprint arXiv:2409.13752}.

\bibitem[{Zhang et~al.(2024{\natexlab{b}})Zhang, Cai, Li, Wu, Hou, and Abdul-Mageed}]{zhang2024distilling}
Chiyu Zhang, Honglong Cai, Yuezhang Li, Yuexin Wu, Le~Hou, and Muhammad Abdul-Mageed. 2024{\natexlab{b}}.
\newblock Distilling text style transfer with self-explanation from llms.
\newblock In \emph{Proc. of NAACL}, pages 200--211.

\bibitem[{Zhang et~al.(2024{\natexlab{c}})Zhang, Yu, Dong, Li, Su, Chu, and Yu}]{zhang2024mm}
Duzhen Zhang, Yahan Yu, Jiahua Dong, Chenxing Li, Dan Su, Chenhui Chu, and Dong Yu. 2024{\natexlab{c}}.
\newblock Mm-llms: Recent advances in multimodal large language models.
\newblock In \emph{Findings of the Association for Computational Linguistics ACL 2024}, pages 12401--12430.

\bibitem[{Zhao et~al.(2024{\natexlab{a}})Zhao, Gan, Wang, Zhou, Yang, Kuang, and Wu}]{zhao2024loraretriever}
Ziyu Zhao, Leilei Gan, Guoyin Wang, Wangchunshu Zhou, Hongxia Yang, Kun Kuang, and Fei Wu. 2024{\natexlab{a}}.
\newblock Loraretriever: Input-aware lora retrieval and composition for mixed tasks in the wild.
\newblock \emph{Proc. of ACL Findings}.

\bibitem[{Zhao et~al.(2024{\natexlab{b}})Zhao, Shen, Zhu, Li, Su, Wang, Kuang, and Wu}]{zhao2024merging}
Ziyu Zhao, Tao Shen, Didi Zhu, Zexi Li, Jing Su, Xuwu Wang, Kun Kuang, and Fei Wu. 2024{\natexlab{b}}.
\newblock Merging loras like playing lego: Pushing the modularity of lora to extremes through rank-wise clustering.
\newblock \emph{arXiv preprint arXiv:2409.16167}.

\bibitem[{Zheng et~al.(2025)Zheng, Shi, Cai, Li, Zhang, Li, Yu, and Ma}]{zheng2025lifelong}
Junhao Zheng, Chengming Shi, Xidi Cai, Qiuke Li, Duzhen Zhang, Chenxing Li, Dong Yu, and Qianli Ma. 2025.
\newblock Lifelong learning of large language model based agents: A roadmap.
\newblock \emph{arXiv preprint arXiv:2501.07278}.

\bibitem[{Zhou et~al.(2023)Zhou, Chen, Wan, Wen, Song, Yu, Huang, Peng, Yang, Xiao et~al.}]{zhou2023characterglm}
Jinfeng Zhou, Zhuang Chen, Dazhen Wan, Bosi Wen, Yi~Song, Jifan Yu, Yongkang Huang, Libiao Peng, Jiaming Yang, Xiyao Xiao, et~al. 2023.
\newblock Characterglm: Customizing chinese conversational ai characters with large language models.
\newblock \emph{arXiv preprint arXiv:2311.16832}.

\end{thebibliography}
\end{document}